\documentclass[conference]{IEEEtran}
\usepackage{times}

\usepackage{times}

\usepackage{multicol}
\usepackage{multirow}
\usepackage{array}
\usepackage{booktabs}
\newcommand{\ra}[1]{\renewcommand{\arraystretch}{#1}}

\usepackage{amsmath, amssymb}
\usepackage{mathtools}
\usepackage{pifont}%
\usepackage[dvipsnames]{xcolor}
\newcommand{\cmark}{\textcolor{ForestGreen}{\ding{51}}}%
\newcommand{\xmark}{\textcolor{Maroon}{\ding{55}}}%
\usepackage{relsize}
\usepackage{dsfont}
\usepackage{graphicx}

\usepackage{subcaption}
\usepackage{leftindex}
\usepackage{xparse}
\NewDocumentCommand{\customVector}{m m m m}{\leftindex^{\mathcal{#2}}{\boldsymbol{#1}}^{\mathcal{#3}}_{#4}}
\NewDocumentCommand{\customVectorFull}{m m m O{}}{\leftindex^{\mathcal{#2}}{\boldsymbol{#1}}^{\mathcal{#3}}_{\mathcal{#4}}}
\newcommand{\norm}[1]{\left\lVert#1\right\rVert}

\usepackage{balance}

\usepackage{todonotes}

\title{Leveling the Playing Field: Carefully Comparing Classical and Learned Controllers for Quadrotor Trajectory Tracking}

\author{\authorblockN{Pratik Kunapuli}
\authorblockA{University of Pennsylvania\\
\href{mailto:pratikk@seas.upenn.edu}{\small \texttt{pratikk@seas.upenn.edu}}}
\and
\authorblockN{Jake Welde}
\authorblockA{University of Pennsylvania\\
\href{mailto:jwelde@seas.upenn.edu}{\small \texttt{jwelde@seas.upenn.edu}}}
\and
\authorblockN{Dinesh Jayaraman}
\authorblockA{University of Pennsylvania\\
\href{mailto:dineshj@seas.upenn.edu}{\small \texttt{dineshj@seas.upenn.edu}}}
\and
\authorblockN{Vijay Kumar}
\authorblockA{University of Pennsylvania\\
\href{mailto:kumar@seas.upenn.edu}{\small \texttt{kumar@seas.upenn.edu}}}
}

\usepackage[numbers]{natbib}
\usepackage{multicol}
\usepackage[bookmarks=true]{hyperref}

\begin{document}

\maketitle

\begin{abstract}
Learning-based control approaches like reinforcement learning (RL) have recently produced a slew of impressive results for tasks like quadrotor trajectory tracking and drone racing.
Naturally, it is common to demonstrate the advantages of these new controllers against established methods like analytical controllers. We observe, however, that reliably comparing the performance of such very different classes of controllers is more complicated than might appear at first sight. 
As a case study, we take up the problem of agile tracking of an end-effector for a quadrotor with a fixed arm.
We develop a set of best practices for synthesizing the best-in-class RL and geometric controllers (GC) for benchmarking. 
In the process, we resolve widespread RL-favoring biases in prior studies that provide asymmetric access to: (1) the task definition, in the form of an objective function, (2) representative datasets, for parameter optimization, and (3) “feedforward” information, describing the desired future trajectory. 
The resulting findings are the following: 
our improvements to the experimental protocol for comparing learned and classical controllers are critical, and each of the above asymmetries can yield misleading conclusions. Prior works have claimed that RL outperforms GC, but we find the gaps between the two controller classes are much smaller than previously published when accounting for symmetric comparisons. 
Geometric control achieves lower steady-state error than RL, while RL has better transient performance, resulting in GC performing better in relatively slow or less agile tasks, but RL performing better when greater agility is required.
Finally, we open-source implementations of geometric and RL controllers for these aerial vehicles, implementing best practices for future development. 
Code, videos, and more can be found on the project website: \href{https://pratikkunapuli.github.io/rl-vs-gc/}{\texttt{https://pratikkunapuli.github.io/rl-vs-gc/}}
\end{abstract}

\IEEEpeerreviewmaketitle

\section{Introduction}\label{sec:intro}
The capabilities of aerial robots have seen explosive growth in recent years, with many exciting results in fast quadrotor flight \cite{bauersfeld2021neurobem}, tracking infeasible trajectories \cite{huang2023datt}, and drone racing, even surpassing human pilots~\cite{kaufmann2023champion}. 
Many of these advances have involved data-driven techniques, and this has spawned careful studies of the design choices in learning-based controller synthesis, such as policy architectures, training procedures, and modeling choices. It is now largely accepted wisdom that data-driven controller synthesis approaches outperform more classical model-based methods for aerial robot control tasks.

\begin{figure}
    \centering
    \includegraphics[width=1.0\linewidth]{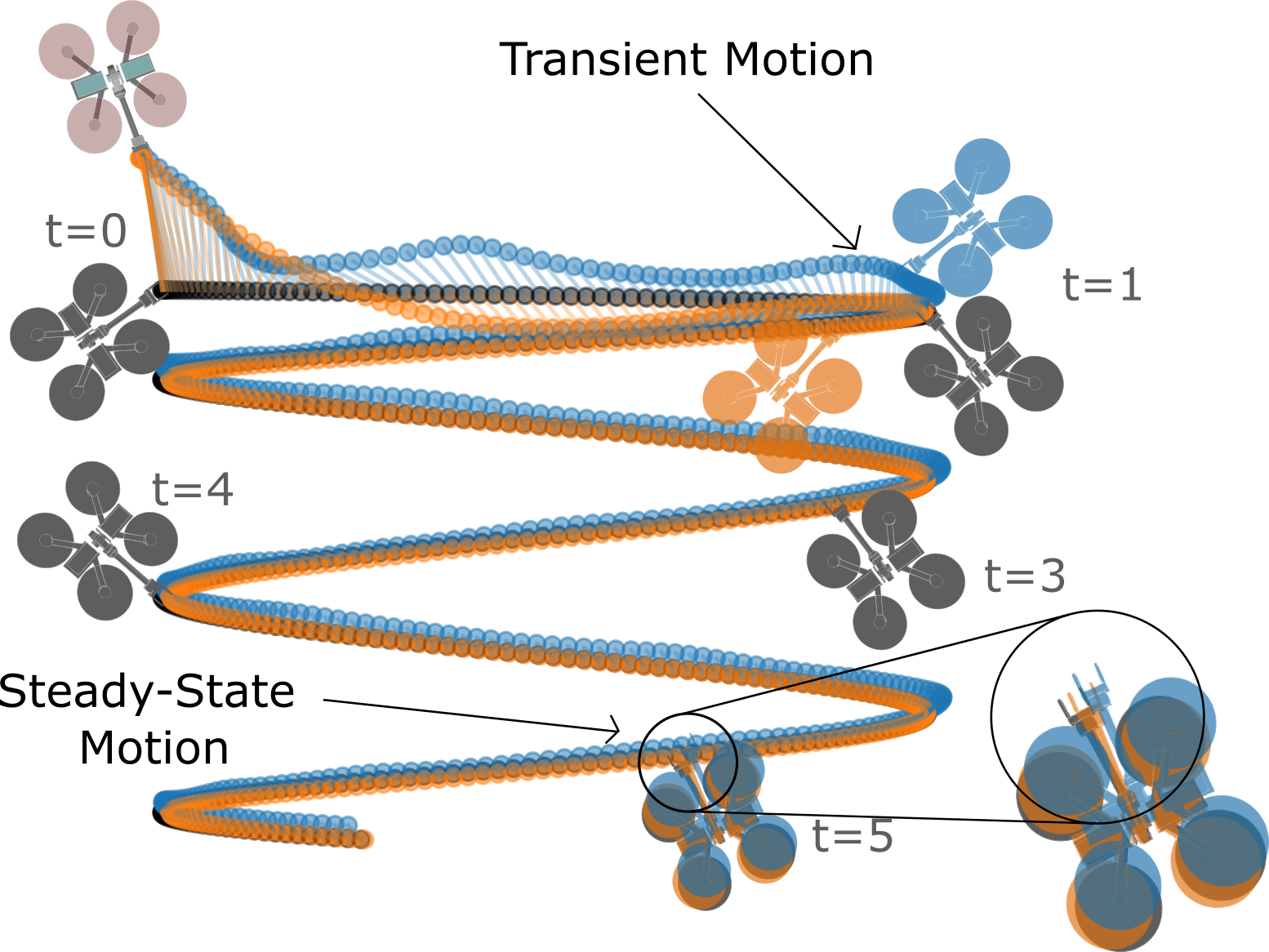}
    \caption{\textbf{Trajectory Tracking for a Quadrotor.} Rollouts of trajectory tracking from an initial perturbation of a reinforcement learning controller (blue) and a geometric controller (orange) on a quadrotor. Robots are visualized at $t=1$ and $t=5$ to highlight transient and steady-state performance relative to the reference trajectory (grey). RL Controller has better transient performance ($t=1$), but worse steady-state error ($t=5$) compared to GC.}
    \label{fig:traj_track}
\end{figure}

However, relatively few studies have systematically compared these two very different classes of controllers. This may be partly be attributed to the small overlap in research communities engaged in the development and application of model-based versus learned controllers. In any case, existing empirical studies largely focus on careful comparisons within each broad model class: such as \citet{sun2024comparativestudynonlinearmpc} who compare model-based controllers, or \citet{kaufmann2022benchmark, dionigi2024power} who benchmark learning-based controllers. 

As we will discuss in detail in Sec~\ref{sec:related} and \ref{sec:methods}, the few prior studies that do compare learned and model-based controllers suffer from some shortcomings. While they aim to reproduce in good faith the respective common practices of the model-based and learned control communities, this inadvertently confounds the comparison, because these standard practices are very different. It is standard in data-driven controller synthesis research to optimize parameters with a carefully designed objective function on a dataset of experiences that exactly match the target task. On the other hand, the tuning of a model-based controller is often a much more heuristic manual procedure to achieve a ``good enough'' configuration. Inheriting such practices when comparing learned and model-based controllers can produce misleading conclusions: the model-based controller may be sub-optimally tuned and perform worse only on that account. 
In other words, correct comparisons across learned and model-based controllers is more complicated than might appear at first sight.

\begin{table*}[t]
    \centering
    \ra{1.2}
    \begin{tabular}{@{}llllllll@{}}
        \toprule
        & & \multicolumn{3}{c}{Geometric Control (GC)} & \multicolumn{3}{c}{Reinforcement Learning (RL)}  \\
        \cmidrule(lr){3-5}  \cmidrule(lr){6-8} 
         Focus & Paper & Tuned for Obj.? & Traj Data? & Feed Forward? & Tuned for Obj.? & Traj Data? & Feed Forward?\\
        \midrule
        \multirow{7}{*}{RL} & Benchmark of Learned Policies \cite{kaufmann2022benchmark} & - & - & - & \cmark & \cmark & \cmark  \\
        & SimpleFlight \cite{chen2024matters}  & - & - & - & \cmark & \cmark & \cmark \\
        & Leveraging Symmetry \cite{welde2024leveraging} & - & - & - & \cmark & \cmark & \cmark \\
        & Sim-2-Multi-Real \cite{molchanov2019sim} & \xmark & \xmark & \xmark & \cmark & \xmark & \xmark \\
        & Power of Input \cite{dionigi2024power} & \xmark & \xmark & \xmark & \cmark & \xmark & \xmark \\
        & DATT \cite{huang2023datt}  & \xmark & \xmark & $\sim$ & \cmark & \cmark & \cmark \\
        & Learning to Fly in Seconds \cite{eschmann2024learning}  & \xmark & \xmark & \xmark & \cmark & \xmark & \xmark \\
        \midrule
        \multirow{4}{*}{GC} & Geometric $SE(3)$ Control \cite{lee2010geometric} & \xmark & \cmark & \cmark & - & - & - \\
        & INDI \cite{tal2021accurate} & \xmark & \cmark & \cmark & - & - & - \\
        & NMPC vs. DFBC \cite{sun2024comparativestudynonlinearmpc} & \xmark & \cmark & \cmark & - & - & - \\
        & PID AutoTune \cite{wang2016automatic} &  \cmark & \cmark & \xmark & - & - & -\\
        & NonLinear PID \cite{moreno2018nonlinear} & \cmark & \cmark & $\sim$ & - & - & - \\
        \midrule
        & Ours & \cmark & \cmark & \cmark & \cmark & \cmark & \cmark \\
        \bottomrule
    \end{tabular}
    \caption{\textbf{Trajectory Tracking Controller Comparisons in Recent Literature.} Partial survey of representative works proposing trajectory tracking methods for quadrotors based on GC and/or RL. We measure these works on whether (to the best of our knowledge) the methods a) optimized a task objective, b) were optimized for trajectory tracking, and c) incorporate future reference information.
    \cmark \thinspace represents a controller had this component, $\sim$ represents a suboptimal implementation (details in Sec \ref{sec:methods}), and \xmark \medspace represents a method that did not have this component. Asymmetries occur when comparing across model classes whose implementation was granted unequal access to task, data, or feedforward information. }
    \label{tab:literature_review}
\end{table*}

In this paper, our \textbf{first contribution} is to identify and fix three key ``asymmetries'' that can produce misleading gains for learning versus model-based methods on standard aerial vehicle tasks like trajectory tracking: objective functions, task-relevant data, and ``feedforward'' inputs specifying upcoming trajectory waypoints. Our \textbf{second contribution} is to apply these proposed improvements in experimental protocol to throughly benchmark reinforcement learned controllers (RL) and geometric controllers (GC) for agile trajectory tracking in quadrotors. Having leveled the playing field, our findings substantially erode the gains of RL versus GC controllers: the two controller classes perform about on par in most of our evaluations, with small gains for GC in steady-state errors, and similar gains for RL in transient performance (Fig \ref{fig:traj_track}). These findings add nuance to today's prevailing wisdom about the relative merits of these controllers. We show further how our improved recipe for empirical comparison can be applied towards validating various hypotheses on new aerial robot classes and tasks. \textbf{Finally,} we provide re-usable implementations of best practices for both RL and geometric controller synthesis for aerial vehicle trajectory tracking tasks including task implementations in a highly parallel RL-ready simulator, to be released publicly together with this paper for future practitioners and researchers.

\section{Related Works}\label{sec:related}
In this section, we elaborate on the deficiencies highlighted in Sec~\ref{sec:intro} in the state of current scientific evidence regarding the efficacy of learned and model-based controllers for quadrotor control.

Comparative studies for quadrotor control have been performed in the past, but a majority of them consider comparisons within the same class of controllers.
Among learning-based methods in quadrotor trajectory tracking tasks, \citet{kaufmann2022benchmark} evaluate the effect of control abstraction on tracking performance in hardware experiments, \citet{dionigi2024power} evaluate the effect of observation on policy performance, and \citet{welde2024leveraging}
evaluate the sample efficiency gains afforded by a symmetry-informed approach to tracking control via RL. 
While comparisons across model-based analytical controllers are fairly straightforward \cite{sun2024comparativestudynonlinearmpc, tal2021accurate}, comparisons involving learning-based methods must contend with a data-driven approach, making it  difficult to perform a correct experimental comparison. As the communities are largely disjoint, it is reasonable that these papers are preoccupied comparing against other works of the same controller class, and thus there are few works that even consider the comparison between RL and analytical controllers like geometric controllers (GC) \cite{mellinger2011design} for trajectory tracking tasks.

Recently, the emerging consensus in the field 
has been that RL-based control is the state-of-the-art for agile trajectory tracking in quadrotors, owing to better computational tractability than Model Predictive Control (MPC) and outperforming GC in tracking error. 
To evaluate such claims,
we summarize the few most relevant prior comparison studies
in Table~\ref{tab:literature_review}. Based on our close reading of these papers, supplementary material, and open-source code (when available), Table~\ref{tab:literature_review} presents the access to task objective functions, task-aligned experiences or datasets, and feedforward waypoint tracking specifications made available to each method considered in these works.
Significant asymmetries are apparent in the access granted to RL and GC methods respectively, which might confound these prior findings --- we show evidence of this in Sec~\ref{sec:experiments}.
Among recent works presenting model-free RL methods, most works train on a proxy \emph{quasi-static} task like hovering\cite{molchanov2019sim, dionigi2024power, eschmann2024learning}, rather than directly on tracking, rendering them likely to perform sub-optimally for agile tracking.
We further find that, of the few works that directly optimize for tracking performance either omit a head-on comparison against an analytical controller \cite{kaufmann2022benchmark, chen2024matters, welde2024leveraging} or compare against an implementation lacking essential components proposed in prior work on analytical control \cite{huang2023datt}, resulting in an asymmetrical comparison. 
Often this is a result of inheriting a baseline controller which may not be optimized for the task at hand, such as an existing module of the firmware of a  commercially-available platform \cite{molchanov2019sim, eschmann2024learning}.
These observations motivate the key contribution of our work: identifying and fixing oversights in these prior experimental comparisons  (Sec \ref{sec:methods}), ultimately enabling us to arrive at more reliable and nuanced conclusions regarding the efficacy of RL and GC controllers for various aerial vehicles and tasks (Sec~\ref{sec:experiments}).

\section{State-of-the-Art Controllers}
In this section we overview two popular classes of controllers, geometric control (GC) and reinforcement learning (RL), that have widely been used for trajectory tracking in quadrotors. Our empirical evaluations and comparisons in later sections will focus on these two classes of controllers.

\subsection{Geometric Control}
Seminal works in the quadrotor control literature \cite{mellinger2011design, lee2010geometric} showed that quadrotors are differentially flat systems and analytically developed a hierarchical controller for this system. This control paradigm uses an explicit control law, constructed from separate position and attitude control loops that are connected via a backstepping-like approach to yield a cascade-like control structure. In this paper, we refer to this controller simply as the ``Geometric Controller" (GC) for brevity (but it is sometimes also referred to as ``$SE(3)$ Control" \cite{lee2010geometric} or ``Differential-Flatness Based Control (DFBC)" \cite{sun2024comparativestudynonlinearmpc}). 

The controller exploits the coupled dynamics between the attitude and position of the system, and controls them with two geometric PD control loops separately. First, a desired linear acceleration is computed from the position PD controller. Next, the desired acceleration (and yaw) is used to compute a desired orientation, which constitutes the reference tracked by the attitude PD controller. The final output of the controller is the collective thrust and moment applied at the center-of-mass (COM) of the vehicle. A detailed implementation of this controller is presented in Appendix \ref{sec:appendix_gc_implementation}.

\subsection{Reinforcement Learning}
Reinforcement learning (RL) seeks to train a policy $\pi_{\theta}$ parameterized by $\theta$, often represented by a neural network. This allows the controller to express a wide range of policies at the expense of requiring data to learn parameters $\theta$. As is standard in RL, the action $a_t$ is produced by passing an observation $o_t$ through the policy according to: $a_t \sim \pi_{\theta}(o_t)$, and the policy is optimized by maximizing the sum of instantaneous rewards $r_t = R(o_t, a_t)$ over time. Policy 
synthesis interleaves experience collection and policy optimization to maximize rewards accumulated over those experiences.
Following previous work including ours~\cite{welde2024leveraging, dionigi2024power}, we define the observation as a \textit{body-frame error representation} concatenating position error $\customVector{e}{B}{}{p} \in \mathbb{R}^3$, orientation error $\customVector{e}{B}{}{R} \in \mathbb{R}^{3\times 3}$, the gravity vector $\customVector{g}{B}{}{} \in \mathbb{R}^3$,  velocity error $\customVector{e}{B}{}{v} \in \mathbb{R}^3$, and angular velocity error $\customVector{e}{B}{}{\omega} \in \mathbb{R}^3$:
\begin{equation}
    o_t = \begin{bmatrix}
        \customVector{e}{B}{}{p} \\
        \customVector{e}{B}{}{R} \\
        \customVector{g}{B}{}{} \\
        \customVector{e}{B}{}{v} \\
        \customVector{e}{B}{}{\omega}                
    \end{bmatrix}
\end{equation}
This representation allows us to use the state of any arbitrary body $\mathcal{B}$ in the observation to the RL policy $\pi_\theta$, given some desired position specified by $\customVector{p}{W}{B}{d}$ and yaw orientation specified by $\customVector{R}{W}{B}{d}$, both expressed in the world frame $\mathcal{W}$. The observation is flattened into a vector in $\mathbb{R}^{21}$. 
The action produced by the policy is $a_t \in \mathbb{R}^4$, clipped to lie within $[-1, 1]$ and then scaled to the collective thrust $f_T$ and body moment $\boldsymbol{M}$ limits of the platform. Further implementation details are provided in Appendix \ref{sec:appendix_rl_implementation}.

\section{Methodology}
\label{sec:methods}
A proper comparative study levels the playing field between methods, eliminates confounding variables, and increases the signal-to-noise ratio from experiments, so that the conclusions drawn offer insights to guide future research and practice. In this section, we identify commonly-overlooked details that yield misleading conclusions in comparative studies for trajectory tracking controllers for aerial vehicles and propose mechanisms to correct them. We finish by summarizing the necessary techniques to perform a proper, fair, and unbiased comparison between learning-based and analytical controllers. 

\subsection{Defining the Trajectory Tracking Task}
The goal of trajectory tracking is to drive an output of the system (\textit{e.g.}, the COM for quadrotors)
asymptotically towards a specified trajectory despite some initial disturbance. To quantitatively compare solutions, we must choose some objective which describes the optimal performance in the task. 

\subsubsection{Trajectories}
Formally, we can define our trajectory as a \emph{time-varying} goal $g(t) \in \mathbb{R}^4$ describing the desired position and yaw of some body in the system.  We sample $g(t)$ at a fixed rate to discretize it into desired waypoints of position and yaw. 

If comparing two methods, they should be optimized on the same tasks, otherwise the training task distribution may be the dominant source of performance differences, confounding the results between the controller classes. Unfortunately, many prior works inadvertently make this error, as seen in Table \ref{tab:literature_review}. 
In hardware experiments, for example, it is common to tune gains for low-level controllers on simple, near-hover tasks but then evaluate the controller with the same gains on any trajectory. This is done primarily for ease of tuning, interpretability, and because of the risk of damaging hardware in case of crashes while tuning on more dynamic trajectories. However, in order to hold controllers equal, all controllers must be optimized or tuned on the same class of tasks for the results to be insightful. 

In this study, we present two versions of trajectory tracking for the quadrotor: \textbf{Hovering} at a specified pose (\textit{i.e.}, tracking a constant reference), and \textbf{Tracking Lissajous curves}, where the desired position and yaw is specified as a Lissajous curve. Mathematically, we represent the trajectories as follows:

\begin{equation}
    \label{eq:Lissajous}
     g_t = \begin{bmatrix}
        x_d \\
        y_d \\
        z_d \\
        \psi_d
    \end{bmatrix} = \begin{bmatrix}
        A_x sin(\omega_x \cdot t + \phi_x) + \delta_x \\
        A_y sin(\omega_y \cdot t + \phi_y) + \delta_y \\
        A_z sin(\omega_z \cdot t + \phi_z) + \delta_z \\ 
        A_{\psi} sin(\omega_{\psi} \cdot t + \phi_{\psi}) + \delta_{\psi}
    \end{bmatrix}
\end{equation}
Hovering is simply a special case of the Lissajous curve, where the amplitudes $A_{x, y, z, \psi}$ are set to 0 and the desired end-effector position and yaw is defined by $\delta_{x, y, z, \psi}$. Details regarding ranges used for randomization in each task is in Appendix \ref{sec:appendix_task}. 

\subsubsection{Initial Conditions}
The distribution from which the initial conditions are sampled is also a core component of the task definition, since even for the same trajectory, varied initial conditions describe vastly different tasked behavior---for example, hovering when initialized at the goal is a very different task than \textit{first} recovering from large initial perturbations and \textit{then} hovering at the goal. In order to hold methods equal, we must explicitly specify the family of initial conditions. Initial conditions for the tasks considered are listed in Table \ref{tab:initial_conditions}. 

\begin{table}[]
    \centering
    \ra{1.2}
    \begin{tabular}{@{}lll@{}}
            \toprule
        & \multicolumn{2}{c}{Task} \\
        \cmidrule(lr){2-3}
        State Component & Hover & Tracking Lissajous\\
        \midrule
        Position ($m$) & [-2, 2] & [-0.5, 0.5] \\
        Velocity ($m/s$)& [0, 0] & [-0.1, 0.1] \\
        Yaw ($rad$) & [-$\pi$, $\pi$] & [-$\pi$, $\pi$] \\
        Angular Velocity ($rad/s$) & [0, 0] & [-0.1, 0.1] \\
        \bottomrule
    \end{tabular}
    \caption{\textbf{Initial Conditions.} Randomization ranges for initialization in Hover and Tracking Lissajous tasks by state component.}
    \label{tab:initial_conditions}
\end{table}

\subsubsection{Objective}
A mathematically specified task objective is necessary to meaningfully compare controllers.
This objective describes \emph{how well} a method accomplishes the task, and serves both to distinguish performance during evaluation as well as provide an objective during optimization and tuning. 
In RL, a simple performance metric such as a ``success rate'' may in theory be optimized directly. However, in practice, such metrics might not offer informative feedback during early stages of training, so it is common practice to add extra ``shaping'' terms to guide RL policy optimization, resulting in a proxy objective~\cite{skalse2022defining,gupta2023behavior}. 
For trajectory tracking, these proxy rewards in prior works commonly resemble an
optimal control objective, like the linear quadratic regulator-style tracking costs used in some works~\cite{kaufmann2022benchmark, molchanov2019sim}.  

We define the objective for the tasks considered following previous work \cite{welde2024leveraging} to measure deviation of the frame of interest from the desired trajectory:
\begin{multline}
\label{eq:rl_reward}
r(t) = \lambda_{p} \phi(\boldsymbol{p}(t) - \boldsymbol{p}_d(t), \delta_{p}) + \lambda_{R} (\norm{\psi(t) - \psi_{d}(t)}) \\
    + \lambda_{v} (\norm{\boldsymbol{v}(t) - \boldsymbol{v}_d(t)}) + \lambda_{\omega} (\norm{\boldsymbol{\omega}(t) - \boldsymbol{\omega}_d(t)}).
\end{multline}
The desired position $\boldsymbol{p}_d(t)$ represents the $x$, $y$, and $z$ coordinates of the desired end-effector waypoint in the world frame. $\boldsymbol{v}_d(t)$ represents the desired velocity, and $\boldsymbol{\omega}_d(t)$ represents the desired angular velocity of the body being tracked in the world frame, both obtained from the derivative of the trajectory $\dot g(t)$. This form follows previous work \cite{rudin2022learning}, with $\phi(\boldsymbol{x}, \delta) := e^{\frac{-\norm{\boldsymbol{x}}}{\delta}}$, allowing the position tolerance to be tuned with $\delta$. The hyperparameters used for the reward function are listed in Table \ref{tab:hyperparameters}.

\begin{table}[]
    \centering
    \ra{1.2}
    \begin{tabular}{@{}llllll@{}}
        \toprule
        Parameter & $\lambda_{p}$ & $\lambda_{R}$ & $\lambda_{v}$ & $\lambda_{\omega}$  & $\delta_{p}$\\
        \midrule
        Value & $15.0 \cdot dt$ & $-4.0 \cdot dt$ & $-0.05 \cdot dt$ & $-0.01 \cdot dt$ & $0.8 \rightarrow 0.1$ \\
        \bottomrule
    \end{tabular}
    \caption{\textbf{Objective Hyperparameters.} Parameters which define the form of the reward function (objective) for trajectory tracking tasks.}
    \label{tab:hyperparameters}
\end{table}

\subsection{Optimizing Controllers}
Here, we present how optimization can be used to choose the respective tunable parameters for each controller class.

\subsubsection{Training the Reinforcement Learning (RL) Policy}
The RL policy is parameterized by a 3-layer MLP with 256 units per layer for the network (details in Appendix \ref{sec:appendix_rl_implementation}). Optimizing the network parameters is done by applying the popular model-free RL method, proximal policy optimization (PPO) \cite{schulman2017proximal}. This approach is standard in the field, where methods optimize a reward function \eqref{eq:rl_reward} over many million simulated environment steps. We found that annealing the position tolerance $\delta_p$ was key to enabling agile behavior while still converging to the goal without large steady-state error. We reduced $\delta_{p}$ by half every 50M timesteps, going from 0.8 at the beginning of training and ending at 0.1 as seen in Table \ref{tab:hyperparameters}.  

\subsubsection{Tuning the Geometric Controller (GC)}
The tunable parameters of the Geometric Controller take the form of PD gains for both the position and attitude loops. Considering each axis separately, this amounts to 12 gains. However, we elect to match the gains for the $\boldsymbol{x}_{\mathcal{V}}$ and $\boldsymbol{y}_{\mathcal{V}}$ axes, owing to the symmetrical control authority over pitch and roll, which is greater than that over yaw ($\boldsymbol{z}_{\mathcal{V}}$). This reduces the number of tunable parameters to 8. 
While hand-tuning this controller is common practice, we propose to use an automatic tuning procedure via Optuna \cite{akiba2019optuna}, which uses Bayesian Optimization to more systematically tune the parameters according to a specified objective function. We use the same reward function as the reinforcement learning controller (Eqn. \ref{eq:rl_reward}), in order to produce the geometric controller tuned with the best performance. 

Tuning gains for the PD controller via optimization is not novel.  
Specific to quadrotor control, \citet{zhu2015gradient} and \citet{cheng2024difftune} presented an approach to tune PD gains using gradient descent, \citet{can2021real} showed how genetic algorithms can be used to tune PID gains, \citet{berkenkamp2016safe} used Gaussian Processes to tune gains, and recently, \citet{zhang2025airpilot} used RL to tune PID gains directly. 
In fact, \citet{loquercio2022autotune} presented an auto-tuning procedure for an MPC controller and found that optimal gains varied even along a single trajectory depending on the motion in various segments. 
In this work, we specifically tune the controller for each task, in the same way that RL is trained on each task, facilitating comparisons between best-in-class controllers. To the best of our knowledge, all of the prior methods comparing RL and GC have used static gains that are specific per-platform for all tasks. %

\subsection{Feedforward Terms}
A core component of differential flatness-based controllers in trajectory tracking is the use of higher-order derivatives of the reference flat outputs (\textit{i.e.}, desired COM position and yaw \cite{faessler2018differential}). The addition of these terms allows the $SE(3)$ controller to choose actions informed by how the reference will move in the future, leading to asymptotic stability even in agile trajectory tracking scenarios. 
These ``feedforward'' terms are a vital component of the GC control laws, and depend on up to $4^{\textrm{th}}$ order derivatives of position and $2^{\textrm{nd}}$ order derivatives of yaw \cite{lee2010geometric, welde2021dynamically}.
Feed forward reference acceleration, $\boldsymbol{\ddot{p}}_d$, appears directly in the computation of the desired acceleration $\ddot{\boldsymbol{p}}_{des}$ in the position PD controller along with the position error, velocity error, and gravitational acceleration. Using standard notation, 
\begin{equation} \label{eq:gc_position_loop}
    \ddot{\boldsymbol{p}}_{des} = -K_p(\boldsymbol{p} - \boldsymbol{p}_d) -K_v(\boldsymbol{v} - \boldsymbol{v}_d) - mg\boldsymbol{z}_{\mathcal{W}} + \ddot{\boldsymbol{p}}_{d}.
\end{equation}
Similarly, the GC computes feedforward angular velocity $\boldsymbol{\omega}_d$ and angular acceleration $\boldsymbol{\dot\omega}_d$ from derivatives of the desired orientation $\boldsymbol{R}_{des}$ (which is determined in part by the desired acceleration $\ddot{\boldsymbol{p}}_{des}$) and uses them in the attitude PD controller: 
\begin{equation}\label{eq:gc_attitude_loop}
\begin{multlined}
    \boldsymbol{\dot\omega}_{des} = -K_R(\boldsymbol{e_R}) -K_{\omega}(\boldsymbol{\omega} - \boldsymbol{\omega}_d) - \\ (\hat{\boldsymbol{\omega}}\boldsymbol{R}^T\boldsymbol{R}_{des}\boldsymbol{\omega}_d - \boldsymbol{R}^T\boldsymbol{R}_{des}\boldsymbol{\dot\omega}_d). 
\end{multlined}
\end{equation}

Commonly, this information is not utilized in GC implementations as seen in Table \ref{tab:literature_review}. Feedforward information in the prior work is often replaced by an integral loop to close steady-state errors \cite{dionigi2024power} in the position control \eqref{eq:gc_position_loop}, or simply omitted \cite{huang2023datt, molchanov2019sim, eschmann2024learning} by setting the feedforward angular velocity and acceleration to 0 in the attitude loop \eqref{eq:gc_attitude_loop}, perhaps due to ease of implementation. This information is crucial to encode the future reference information, and contributes a significant performance gain, as we will verify in Sec~\ref{sec:experiments}. 

In fact, the benefit of leveraging future information is also seen in many recent reinforcement learning controllers for trajectory tracking \cite{kaufmann2022benchmark, huang2023datt}. Often, this information appears as a horizon of future waypoints or goals appended to the observation. In order to give both controllers access to the same information and to avoid privileging one controller over the other, we append a sequence of future position and yaws to the observation for both the GC and RL controllers. The horizon length used is $H=10$, corresponding to samples from $t+dt$ to $t+10dt$ of the reference trajectory. This horizon is sufficiently large for the GC controller to approximate high-order derivatives of the reference via finite differencing, while the RL controller directly observes the horizon without imposing any structure \textit{a priori}.

\subsection{Summing Up: Best Practices For Benchmarking}
In summary, there are a number of advantages that could be afforded to methods of either class, but have often been inadvertently only presented to RL in standard implementations. As RL is a data-driven approach, it must use some data and objective function, and the fair practice to advantage GC equally is to use a tuning algorithm on the same objective and 
data access. 
Furthermore, when specifically considering trajectory tracking, future reference information is a core component of the geometric controller, and both RL and GC should get the same access to future reference information in the form of a horizon of waypoints. Holding equal the optimization objective, data access, and future reference access is imperative to properly evaluate differences in controller synthesis approaches.

Prior works have compared RL controllers against hand-tuned GC implementations (asymmetric optimization), which may have been tuned for hovering (asymmetric data access). Even trajectory tracking works like \cite{huang2023datt} compare against a controller using PID (asymmetric access to future reference information), and do not use feedforward information equally between the RL and GC method compared. Asymmetries in prior comparisons are summarized in Table \ref{tab:literature_review}.

\section{Experimental Results} 
\label{sec:experiments}

In this section, we seek to first validate the methodology proposed for fairly comparing learning-based and analytical controllers, demonstrating how correcting for such asymmetries leads to strengthened baseline models. Next, armed with 
this improved
experimental protocol to compare methods fairly, we seek to answer specific questions about trajectory tracking of aerial vehicles.

\begin{figure*}[t]
    \centering
    \includegraphics[width=0.9\linewidth]{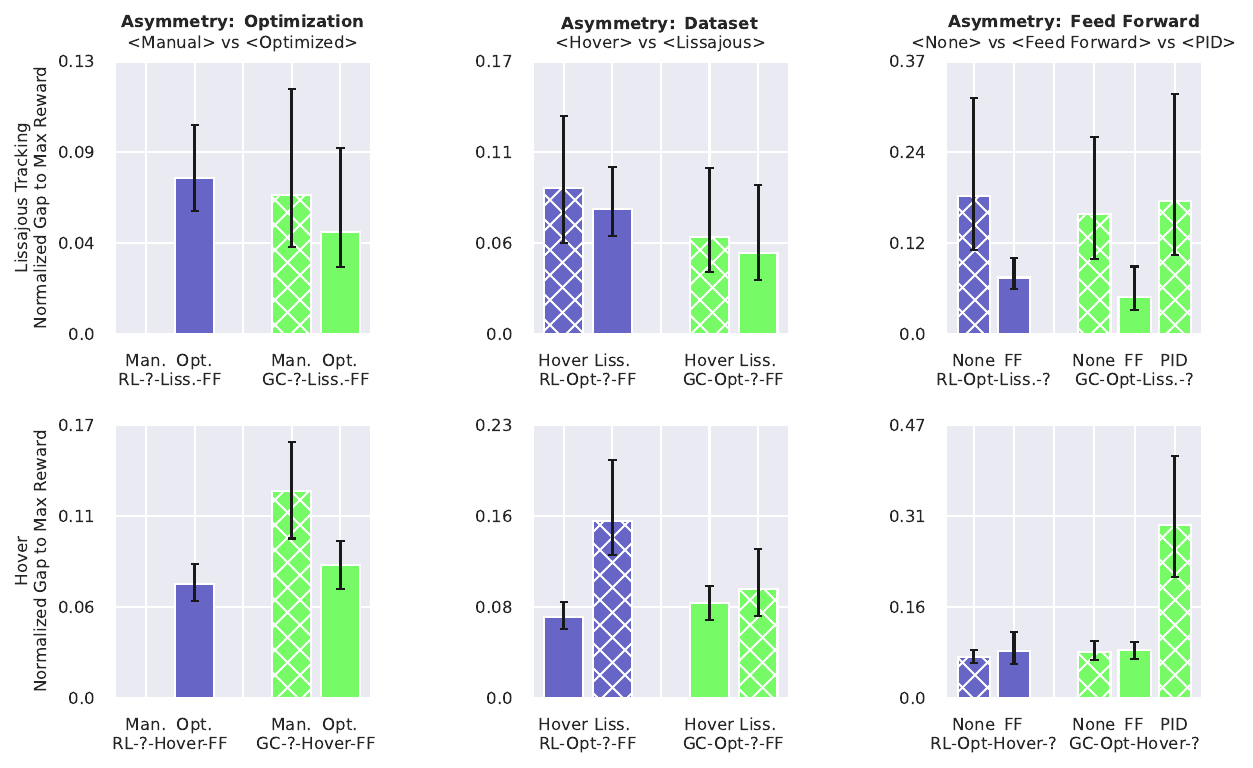}
    \caption{\textbf{The Impact of Each Type of Asymmetry in RL vs.~GC Comparisons.} Model comparisons in both Lissajous Tracking (top row) and Hover (bottom row) tasks for the aerial manipulator vehicle. Results are shown with median and inter-quartile range across 1000 evaluations per task. Controllers are measured on normalized gap to maximum reward to highlight improved performance by correcting asymmetries. Hatched bars represent controller variants that are a suboptimal choice within each category. Comparisons are isolated into asymmetric access to optimization (left), asymmetric access to data (middle), and asymmetric access to future reference information (right). All controllers improve by conducting the protocols presented for optimization, data access, and use of feed forward information, but this is seen to make a larger difference in the Lissajous Tracking task.}
    \label{fig:model_ablations}
\end{figure*}

\subsection{Simulation Environment} \label{sec:environment_details}
We perform the experiments in simulation, giving us the best ability to hold methods equal in data access, initial conditions, and optimization, and to evaluate over large numbers of trials. 
To simulate the aerial vehicles during training and evaluation of policies, we implement the dynamics in IsaacLab \cite{mittal2023orbit}, using the IsaacSim \cite{liang2018gpu} physics engine. Following prior work leveraging massively parallelized simulation \cite{rudin2022learning}, the environment is implemented with GPU-level parallelism, allowing for a large number of simultaneous simulations. We ensure the same goal configurations are randomly sampled across evaluated methods, regardless of control strategy selected. Following prior work \cite{kaufmann2022benchmark}, we simulate the dynamics at $100Hz$, and the controllers are run at $50Hz$. Training is performed on 4096 simultaneous environments over 200 million total timesteps in approximately 30 minutes ($\sim175,000$ steps/sec) on an NVIDIA RTX A5000 GPU. The code to run this environment (as well as training scripts) is open-sourced in order to accelerate the development of agile aerial manipulation. 

In this study, we consider a quadrotor platform with a rigidly attached arm. This ``aerial manipulator'' morphology is a generalization of the quadrotor, and serves as a platform to evaluate new hypotheses regarding control of aerial vehicles. We expect that the GC will suffer in this morphology since it is principally designed to operate on the COM as opposed to controlling the end-effector directly, whereas RL can be applied to any body in the system. Together with the quadrotor, this aerial manipulator robot allows us to answer the questions posed above.

\subsection{Validating the Experimental Protocol}

In this section, we show how previous methods may have drawn conclusions between RL-based methods and GC-based implementations that bias towards better results for the RL-based controller. Specifically, we show how correcting the asymmetries laid out in the methodology (access to the optimization objective, access to the dataset, and use of future reference information) lead to significantly improved performance of both controllers, closing the the perceived gap in controller performance in recent literature. We construct models from the controller type: \{RL, GC\}, optimization strategy: \{Manual (Man.), Optimized (Opt.)\}, dataset: \{Hover (Hov.), Lissajous (Liss.)\}, and feed forward strategy: \{None, Feedforward (FF), Integral (PID)\}. We optimize (train) every method according to the dataset and optimization procedure described, and evaluate the models in 1000 rollouts in the Lissajous Tracking and Hover tasks, highlighting the impact of various choices of optimization, data access, and feed forward information in Fig \ref{fig:model_ablations} for the aerial manipulator morphology. A full presentation of the achieved reward, position tracking RMSE, and yaw tracking RMSE for all models is left to the supplementary material. Fig \ref{fig:model_ablations} plots how much short of maximum achievable reward each method falls on average, to highlight the impact of removing asymmetries.

\subsubsection{Optimizing Controllers}
First, we consider the effect of optimizing the controller in the left column of Fig \ref{fig:model_ablations}. In these experiments, we compare how controllers benefit from optimizing based on some objective. Note, that it is not feasible to ``hand-tune'' the neural network RL controllers we study in this work, and thus we omit this variant. Comparatively, we can observe how tuning for an objective affects the GC, and we see that tuning the controller for the objective lowers the gap to maximum reward, improving performance. In this experiment, we allowed the methods to optimize parameters from the measured task (Lissajous tracking in top row, Hover in bottom row), and utilize feedforward information so this represents the best-performing model. 

\subsubsection{Data Access}
In the middle column of Fig \ref{fig:model_ablations}, we compare the effect of task-aligned experience during optimization on the controller performance. We see that models optimized on another task (Hover for the Lissajous Tracking evaluation and vice versa) perform worse than those optimized for the same task as that on which they are evaluated. This result is expected for RL (since it is a data-driven method), but note that GC also benefits from this practice, which should thus be adopted. Commonly, GC gains are hand-tuned in \textit{quasi-static} motions and set for all downstream tasks like trajectory tracking, and we can see that this practice is inherently disadvantaging the GC. 

\subsubsection{Future Reference Information}
Finally, in the last column (Fig \ref{fig:model_ablations}) we investigate the role of feedforward information for both Lissajous Tracking and Hover. In the top row we can clearly see that for Lissajous tracking both RL and GC benefit from the addition of future reference information. Interestingly, we find that PID as a substitute of feedforward information in GC performs worse than if no feedforward information was provided at all, perhaps due to the integral terms being sensitive to the gains which are auto-tuned. In the bottom row, we find that utilizing feedforward information presents no benefit due to the static nature of the reference in the Hover task, and all models are nearly perfect in this task, except for the GC controller using PID. 

\subsubsection{Conclusions}
From these results, it is evident the best performing GC controller is the one which optimizes a reward, has access to the same dataset being evaluated on, and uses feedforward terms (GC-Opt.-Liss.-FF). This policy adopts the strategies proposed in this work to reduce asymmetries in the three areas, and not performing these corrective measures in any one category increases the gap to optimal reward, creating a suboptimal policy. Indeed, we can examine how previous RL methods had implemented the analytical baseline, and show why their claims of superior performance over the baseline may have been misguided or overstated. For trajectory tracking, comparing against a GC controller hand-tuned in hovering (GC-Man.-Hover-None) \cite{molchanov2019sim}, optimizing for the wrong dataset (GC-Opt.-Hover-None), or using PID instead of feedforward terms (GC-Opt.-Liss.-PID or GC-Opt.-Hover-PID) \cite{huang2023datt, dionigi2024power, eschmann2024learning} all result in misguided comparisons since the baseline method is handicapped. This ultimately results in an over-estimate of the performance gap between RL and GC in trajectory tracking, which cannot be attributed to the method itself as previously thought. We open-source the implementation and optimization code for the strengthened controllers, hoping to accelerate research in agile aerial vehicles. We hope to impress the importance of these corrections of model-based controllers to the research community to improve baselines and draw more robust conclusions.

\subsection{Best-of-the-Best Comparison}

\begin{figure}
    \centering
    \includegraphics[width=\linewidth]{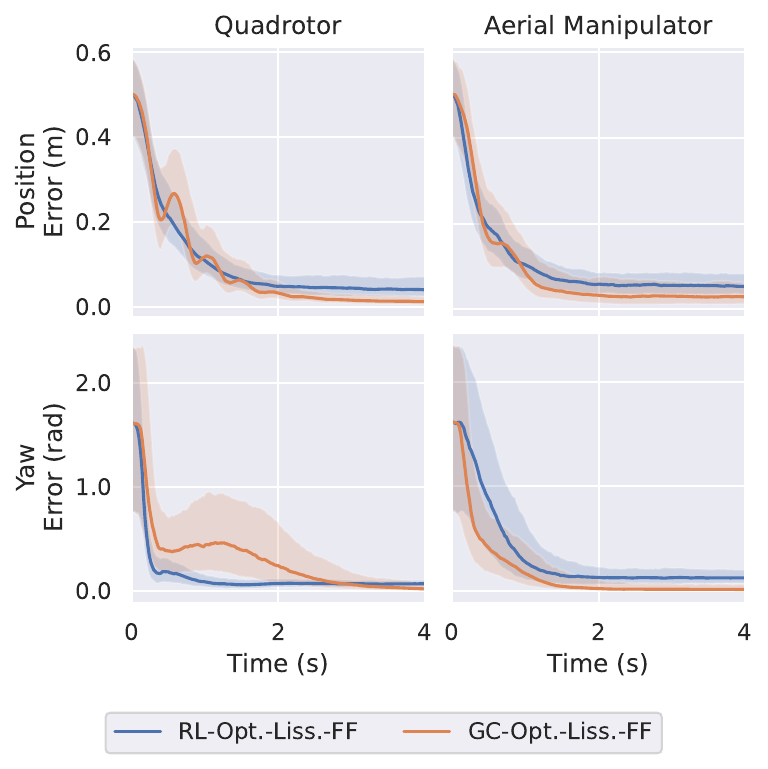}
    \caption{\textbf{Trajectory Tracking Errors.} Position and yaw errors over time for both the best-in-class RL and GC controllers evaluated on the Quadrotor and Aerial Manipulator morphologies. Results are shown as the median with inter-quartile range shaded from 1000 evaluations per morphology in Lissajous Tracking.}
    \label{fig:trajectory_tracking}
\end{figure}

\begin{table*}[t]
    \centering
    \ra{1.2}
    \begin{tabular}{@{}l l l l l l l @{}}
        \toprule
        & \multicolumn{3}{c}{Quadrotor} & \multicolumn{3}{c}{Aerial Manipulator} \\
        \cmidrule(lr){2-4} \cmidrule(lr){5-7} 
        Controller & Avg. Reward & Position RMSE (m) & Yaw RMSE (rad) & Avg. Reward & Position RMSE (m) & Yaw RMSE (rad)  \\
        \midrule
        RL-Opt.-Liss.-FF & \textbf{14.196 $\pm$ 0.48} & \textbf{0.119 $\pm$ 0.05} & \textbf{0.274 $\pm$ 0.15} & 13.621 $\pm$ 1.28 &\textbf{0.118 $\pm$ 0.05} & 0.487 $\pm$ 0.26 \\
        GC-Opt.-Liss.-FF & 13.447 $\pm$ 1.61 & 0.158 $\pm$ 0.20 & 0.483 $\pm$ 0.29 & \textbf{13.792 $\pm$ 1.28} & 0.136 $\pm$ 0.10 & \textbf{0.405 $\pm$ 0.29} \\
        \bottomrule        
    \end{tabular}
    \caption{\textbf{Trajectory Tracking for Quadrotor and Aerial Manipulator.} Comparison of the best RL controller against the best GC controller in trajectory tracking for a Quadrotor and an Aerial Manipulator (Quadrotor with fixed-arm) over 1000 trials in the Lissajous Tracking task. Rewards are averaged over time per rollout, then presented as average and standard deviation over the trials. Maximum reward is $15.0$. RMSE of position and yaw are shown as averages with standard deviation over 1000 trials. Contrary to many literature claims, performance is very similar between best-in-class methods.}
    \label{tab:steel_man_rmse}
\end{table*}
By using this protocol to obtain best-in-class methods for both RL and GC controllers, we can now perform experiments to develop controllers for trajectory tracking of the quadrotor and the aerial manipulator. We seek to properly evaluate the claim that RL outperforms GC for trajectory tracking. We develop the best-in-class controllers for both morphologies using both RL and GC, and evaluate the robots on Lissajous Tracking in 1000 evaluations, showing the median error over time with inter-quartile range for both position and yaw tracking in Fig. \ref{fig:trajectory_tracking}. Root-mean-square-error (RMSE) for these experiments are recorded in Table \ref{tab:steel_man_rmse}. 

Both controllers work well to quickly reduce the error in position and yaw of large initial perturbations. Although the RMSE metrics seem to indicate the RL controller has a slight edge over the GC controller, we see in Fig. \ref{fig:trajectory_tracking} that the GC error converges to 0 while the RL controller has a slight steady-state offset. While RMSE is a standard metric to report in trajectory tracking, we notice it is sensitive to the scale of the inital perturbation and the duration of the episode (where long episodes emphasize steady-state error and short episodes emphasize transient performance), and as such may not reflect the overall or asymptotic performance of the controller for some downstream task. In summary, although the best-in-class RL controller achieves lower RMSE in trajectory tracking for both a quadrotor and aerial manipulator compared to the best-in-class GC controller, the steady-state error does not converge to zero. This suggests that the RL controller is better at achieving transient performance, while the GC achieves better asymptotic performance at the cost of near-term error, echoing results shown in \cite{eschmann2024learning}. 
Thus, we conclude that GC outperforms RL on the reward objective and RL only has a slight advantage in position RMSE from large perturbations due to better initial transient performance, reversing claims of prior literature. 

Additionally, we see that from these results both controllers work well to control the desired reference point, whether it is the COM location in the quadrotor or the end-effector in the aerial manipulator. This shows impressive results for the GC, where observing and controlling the COM allows excellent performance in the end-effector tracking although the controller is not formulated exactly for this case.

\subsection{Does RL beat GC?}
In order to revaluate claims from prior works (Table \ref{tab:literature_review}) of RL beating GC, we choose to evaluate on the Hover task, which isolates the transient component of the trajectory tracking task (since the reference trajectory is stationary). In this setting, we can evaluate how the controllers can reduce their position and yaw errors as a function of time from a large initial perturbation. As seen in Fig \ref{fig:model_ablations}, for the Hover task, the feed-forward terms do not yield any significant benefit in performance due to the stationary trajectory. Thus, we use the RL-Opt.-Hover-None as the RL controller, and GC-Opt.-Hover-None as the GC controller. Additionally, since we evaluate this on a quadrotor with a fixed arm (that is, a 0-DOF Aerial Manipulator \cite{welde2021dynamically}), we can directly compare controller assumptions by presenting an RL controller observing the end-effector state (RL-EE), a GC controller which can only by design observe the COM state, and an RL controller observing the same COM state as the GC (RL-COM).  This comparison allows us to directly investigate if observing only the COM state is inherently limiting for control of the end-effector. The insights from this experiment are only possible due to the experimental protocol optimizing the controllers in the same way, isolating resultant differences in performance to the inherent controllers. 

\begin{figure} [h]
    \centering
    \includegraphics[width=0.9\linewidth]{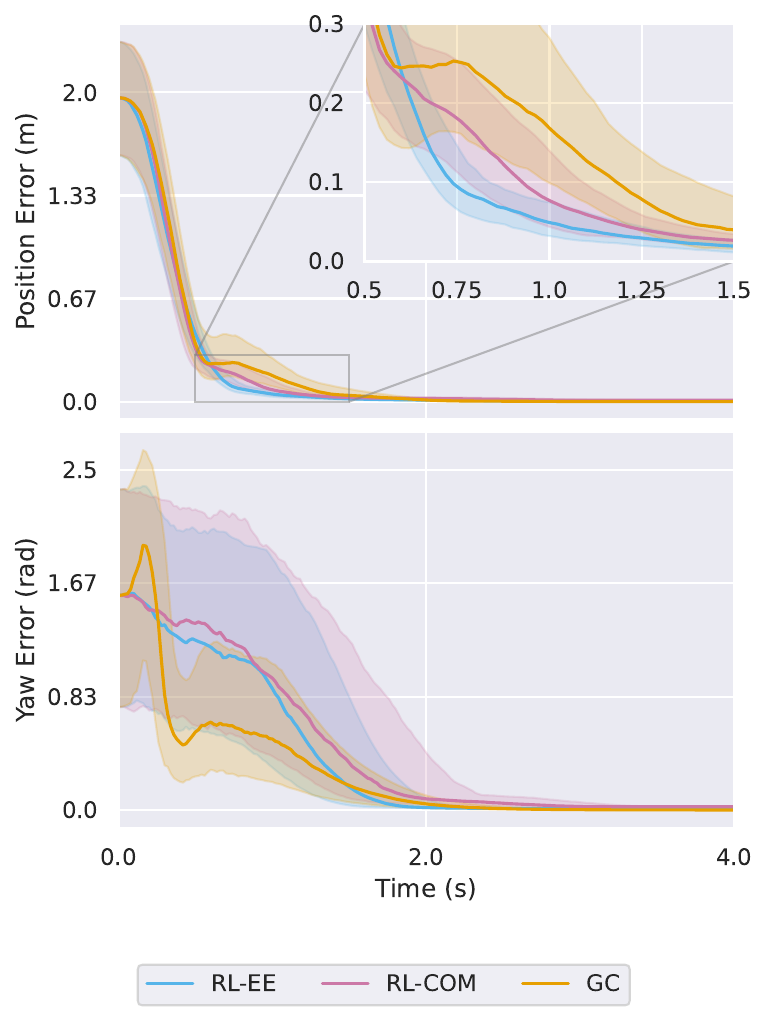}
    \caption{\textbf{Hover Errors.} Position and yaw errors over time for the RL controller observing the end-effector (RL-EE), RL controller observing the COM (RL-COM), and the geometric controller (GC). Errors are shown as the median with inter-quartile ranges across 1000 evaluations in the Hover task.}
    \label{fig:hover_errors}
\end{figure}

We evaluate this for 1000 random goal positions, and present the results in Fig. \ref{fig:hover_errors}. This evaluation is also motivated by the real-world task of catching projectiles (Ball-Catching), for which agility and transient performance are essential. 
Thus, we present an evaluation scenario in which a ball is thrown with some random initial velocity, and the quadrotor aims to catch the ball at a given height off the ground. Time-to-Catch is a metric used to roughly define difficulty of the task, where higher times correspond to easier tasks and lower times correspond to more difficult tasks. This is modulated by affecting the vertical component of the initial velocity which moves the catch location as well as the time to reach the catch plane. 
We present the catch success percentages in Table \ref{tab:ball_catching_results} for 100 evaluations of 5 catch opportunities each, with varied initial velocities corresponding to different time allocations for the catching. Please view videos of this task in the supplementary materials.

\begin{table}[]
    \centering
    \ra{1.2}
    \begin{tabular}{@{}l l l l l @{}}
        \toprule
          & \multicolumn{4}{c}{Time-To-Catch (s)} \\
         \cmidrule(lr){2-5} 
         Controller & 0.79 & 1.09 & 1.53 & 1.99 \\
         \midrule 
         RL-EE & 0.65  & \textbf{1.0} & \textbf{1.0} & \textbf{0.99 }\\
         RL-COM & \textbf{0.72} & 0.75  & 0.85  & 0.94  \\
         GC & 0.30  & 0.37  & 0.49  & 0.97  \\
         \bottomrule
    \end{tabular}
    \caption{\textbf{Ball Catching success rate.} Percentage of catches made by the aerial manipulator with varied initial velocities by each controller, denoted by the time-to-catch in each setting. Results are shown as mean catch rate over 100 trials.}
    \label{tab:ball_catching_results}
\end{table}

Here we can see that there is a manifest difference between the controllers, where RL-EE is shown to have the best transient performance in reducing position error compared to the other controllers, resulting in the highest success percentages of balls caught. The GC controller, although guaranteed to converge to 0 steady-state error, has the worst transient performance among the controllers, resulting in poor performance in the ball catching task. The RL-COM performance indicates that the COM state observation is not the impeding factor for the GC transient performance, as the RL-COM has better transients, but rather the structured hierarchical control of the GC may be limiting for very agile behavior, especially of an end-effector. 

\subsection{Data Asymmetry under Domain Randomization}
Thus far, we have considered data asymmetry as access in simulation to rollout data under a particular task. 
Another use of simulation data is to perform \emph{domain randomization}, a common technique used to bridge the simulation to real-world gap by simulating a distribution of parameters that may be unknown or hard to model. 
Often, this is used with RL in order to train a single model that is able to generalize or perform well on any test-time parameters that fall within the distribution. 
However, domain randomization can also be applied to GC, and in alignment with evaluating the controllers fairly, we train the quadrotor trajectory tracking task under different ranges of domain randomization varying the mass, inertia tensor, and thrust-to-weight ratio of the vehicles. 
The training distributions are 0\%, i.e. no domain randomization (RL-0, GC-0), 20\% (RL-20, GC-20), and 40\% (RL-40, GC-40) uniformly sampled centered at the nominal value for each mass, inertia, and thrust-to-weight. 
The evaluation is performed on 1000 random samples from the 20\% distribution. All controllers evaluated here perform optimization, are trained on the Lissajous task, and use feed forward information, matching best practices established in this work (i.e. RL-Opt-Liss-FF and GC-Opt-Liss-FF). 
Results can be seen in Table \ref{tab:domain_randomization_results}. 

\begin{table}[]
    \centering
    \ra{1.2}
    \begin{tabular}{@{}l l l l @{}}
        \toprule
        Controller & Avg. Reward & Position RMSE (m) & Yaw RMSE (rad)\\
        \midrule 
        RL-0 & 13.345 $\pm$ 1.30 & 0.124 $\pm$ 0.07 & 0.281 $\pm$ 0.18 \\
        RL-20 & \textbf{13.558 $\pm$ 1.08} & 0.119 $\pm$ 0.06 & \textbf{0.260 $\pm$ 0.16} \\
        RL-40 & 13.506 $\pm$ 1.11 & \textbf{0.113 $\pm$ 0.06} & 0.301 $\pm$ 0.18 \\
        \midrule
        GC-0 & 12.005 $\pm$ 2.14 & 0.163 $\pm$ 0.20 & 0.461 $\pm$ 0.28 \\
        GC-20 & 12.162 $\pm$ 2.47 & 0.161 $\pm$ 0.25 & 0.485 $\pm$ 0.31 \\
        GC-40 & 11.834 $\pm$ 2.74 & 0.216 $\pm$ 0.33 & 0.510 $\pm$ 0.30 \\
        \bottomrule
    \end{tabular}
    \caption{\textbf{Controller Performance under Domain Randomization.} Comparison of RL and GC controllers optimized for Lissajous trajectory tracking, with feed-forward terms, under varied amounts of domain randomization (0-40\%) of mass, inertia, and thrust-to-weight. Results are shown as averages with standard deviation over 1000 trials in the 20\% domain randomization setting for evaluation.}
    \label{tab:domain_randomization_results}
\end{table}
All controllers in this setting perform worse than without domain randomization, owing to the increased difficulty of the test-time task. The RL controllers exhibit less degradation, matching the low position and yaw RMSE from the evaluation setting without domain randomization, but perform slightly worse on average reward (seemingly due to angular velocity oscillations). The GC performs significantly worse than before, largely due to the model-based nature of the controller,  since the relatively small number of optimizable parameters are not able to correct for the model uncertainty at evaluation time. Here we can see that RL is more amenable to the domain randomization technique for sim-to-real transfer, and can be a preferred controller in settings where simulation is needed to overcome uncertainty in test-time configurations. 

\subsection{Realistic Dynamics}
\begin{table}[t]
    \centering
    \ra{1.2}
    \begin{tabular}{@{}l l l l @{}}
        \toprule
        Controller & Avg. Reward & Position RMSE (m) & Yaw RMSE (rad)\\
        \midrule 
        RL-Simple & 2.941 $\pm$ 1.69 & 0.856 $\pm$ 0.20 & 1.695 $\pm$ 0.17 \\
        RL-Realistic & \textbf{13.053 $\pm$ 1.17} & \textbf{0.164 $\pm$ 0.06} & \textbf{0.236 $\pm$ 0.18} \\
        \midrule
        GC-Simple & 12.709 $\pm$ 0.72 & 0.234 $\pm$ 0.12 & 0.595 $\pm$ 0.18 \\
        GC-Realistic & 12.595 $\pm$ 1.99 & 0.183 $\pm$ 0.28 & 0.406 $\pm$ 0.31 \\
        \bottomrule
    \end{tabular}
    \caption{\textbf{Controller Performance under Realistic Dynamics.} Comparison of RL and GC controllers optimized for Lissajous trajectory tracking, with feed-forward terms, under Simple dynamics (rigid body dynamics only) or Realistic dynamics (motor dynamics and saturation). Results are shown as averages with standard deviation over 1000 trials in the realistic dynamics setting for evaluation.}
    \label{tab:dynamics_ablation_results}
\end{table}
Often, quadrotor dynamics are simulated as a simple rigid body system (as in this work). 
In reality, there are more complex dynamics involved in the real-world counterpart of the dynamics, including but not limited to: motor delay, actuator saturation, communication delay, etc. 
Many of these effects can be mitigated by running tight control loops at a fast rate on-device, as opposed to relying on an off-board controller. 
However, since some of these complex dynamics can be modeled, a question arises as to whether controllers optimized in simulation can take advantage of these phenomena for better test-time performance. 
Specifically, we aim to answer whether RL can learn from these realistic dynamics more than GC, since RL is a data-driven method whereas GC only considers the rigid body dynamics. 
We trained models with and without the realistic dynamics (details in Appendix \ref{sec:appendix_motor_dynamics}) for both RL (RL-Simple, RL-Realistic) and GC (GC-Simple, GC-Complex) and evaluate on settings where motor dynamics are modeled as a first-order system and motor saturation and allocation is considered, but controller delay is assumed to be 0 from an on-board controller.

We present the results of this ablation in Table \ref{tab:dynamics_ablation_results}. We find that the RL model trained in the simple dynamics fails to perform well under the realistic dynamics, likely due to over-fitting to the simple dynamics and performing ``bang-bang'' control, which is not feasible with respect to the motor dynamics. The best-performing model, RL-Realistic, is trained and evaluated under the more complex dynamics and is able to learn under the improved dynamics, as expected. We expect this model to be able to transfer most reliably to a real-world platform. 
Comparatively, the GC controllers do not perform as well, owing to the simplicity of the dynamics model considered in the controller. 
However, the change from simple to realistic dynamics does not affect performance nearly as much as it does for the RL controllers, and both GC-Simple and GC-Realistic are able to perform similarly, albeit slightly worse than the best-performing data-driven method. 
Ultimately, we see that using realistic dynamics models improves the expected transferability of the RL methods due to the data-driven nature of the controller, but does not affect GC controllers in the same way.
We expect that both the GC controllers and the RL-Realistic controller will be able to transfer from simulation to a real robot, and the controller performance will be similar to results presented here in simulation. 

\section{Discussion and Limitations}
With the robust protocol established in this work, we are able to correct asymmetries in comparisons from prior work, revealing new insights into controller performance for quadrotor trajectory tracking. We show that reinforcement learning (RL) controllers do not always perform better than geometric control (GC), and only perform better in specific transient settings at the expense of steady-state error. For some particularly agile tasks like ball-catching, this results in significantly improved performance, but for tasks requiring less agility the GC performs better. Additionally, we are able to evaluate controller methods under new hypotheses and since we ensure models are fairly optimized, we can attribute results directly to the controller. This experimental protocol should be used for any comparison involving data-driven controllers being compared against model-based controllers, as not carefully applying the same advantages can mislead conclusions as we have shown in our experiments. 

One of the main limitations of this work is that the evaluations are performed in simulation as opposed to on real hardware, which would represent the ultimate desired testing condition. 
We perform experiments under domain randomization, with accurate first-order motor dynamics, and under control delay to simulate real-world phenomena that affect performance of simulation policies during transfer to hardware. These protocols closely follow other successful sim2real works \cite{zhang2023learning, eschmann2024learning} and thus inspire confidence that the policies will retain performance outside of simulation.  
We also do not reproduce the exact experiments of previous works in terms of modeling the same training distributions, model architecture innovations, or evaluation platforms. We instead propose a thorough and principled approach to benchmark controllers of various classes fairly, and leave it for future work to reproduce the experiments of earlier works under this experimental protocol to dissect contributions. 
Finally, it is also left to future work to translate our prescribed evaluation procedures to other robot morphologies and tasks, in order to ascertain whether the conclusions obtained here in trajectory tracking problems for quadrotors and fixed-arm aerial manipulators also persist more broadly.

\section{Conclusions} 
In this work, we present a robust experimental protocol correcting common asymmetries in comparing RL and GC controllers, as well as how failing to account for these asymmetries yield misleading conclusions. We execute this protocol in order to study trajectory tracking performance for end-effector control of a fixed-arm underactuated aerial manipulator, and contextualize results for the quadrotor setting. We show that the RL controller performs better in the transient setting at the expense of steady-state error, and that the gaps between the two controllers are very close when both are the best-in-class versions of the controllers. We demonstrate a practical scenario in which transient performance matters via a ball-catching task and show that the RL controller is able to out-perform the GC in this setting. We also point to common approaches to generalize sim2real behavior and show that our conclusions hold under these settings. We believe this work will guide future researchers in developing more agile aerial manipulators, and elucidate better comparisons between learning-based and classical controllers. 

\section{Acknowledgments}
We gratefully acknowledge the support of ARL DCIST CRA W911NF-17-2-0181, NSF Grant CCR-2112665, NSF CAREER Award 2239301, DARPA TIAMAT HR00112490421, ONR award N00014-22-12677, NSF Grant 331783, NIFA grant 2022-67021-36856, the IoT4Ag Engineering Research Center funded by the National Science Foundation (NSF) under NSF Cooperative Agreement Number EEC-1941529, and NVIDIA.

\balance 
\bibliographystyle{plainnat}
\bibliography{references}

\begin{thebibliography}{32}
\providecommand{\natexlab}[1]{#1}
\providecommand{\url}[1]{\texttt{#1}}
\expandafter\ifx\csname urlstyle\endcsname\relax
  \providecommand{\doi}[1]{doi: #1}\else
  \providecommand{\doi}{doi: \begingroup \urlstyle{rm}\Url}\fi

\bibitem[Akiba et~al.(2019)Akiba, Sano, Yanase, Ohta, and Koyama]{akiba2019optuna}
Takuya Akiba, Shotaro Sano, Toshihiko Yanase, Takeru Ohta, and Masanori Koyama.
\newblock Optuna: A next-generation hyperparameter optimization framework.
\newblock In \emph{Proceedings of the 25th ACM SIGKDD international conference on knowledge discovery \& data mining}, pages 2623--2631, 2019.

\bibitem[Bauersfeld* et~al.(2021)Bauersfeld*, Kaufmann*, Foehn, Sun, and Scaramuzza]{bauersfeld2021neurobem}
Leonard Bauersfeld*, Elia Kaufmann*, Philipp Foehn, Sihao Sun, and Davide Scaramuzza.
\newblock Neurobem: Hybrid aerodynamic quadrotor model.
\newblock In \emph{Robotics: Science and Systems XVII}, RSS2021. Robotics: Science and Systems Foundation, July 2021.
\newblock \doi{10.15607/rss.2021.xvii.042}.
\newblock URL \url{http://dx.doi.org/10.15607/RSS.2021.XVII.042}.

\bibitem[Berkenkamp et~al.(2016)Berkenkamp, Schoellig, and Krause]{berkenkamp2016safe}
Felix Berkenkamp, Angela~P Schoellig, and Andreas Krause.
\newblock Safe controller optimization for quadrotors with gaussian processes.
\newblock In \emph{2016 IEEE international conference on robotics and automation (ICRA)}, pages 491--496. IEEE, 2016.

\bibitem[Can and Ercan(2021)]{can2021real}
Muharrem~Selim Can and Hamdi Ercan.
\newblock Real-time tuning of pid controller based on optimization algorithms for a quadrotor.
\newblock \emph{Aircraft Engineering and Aerospace Technology}, 94\penalty0 (3):\penalty0 418--430, 2021.

\bibitem[Chen et~al.(2024)Chen, Yu, Xie, Gao, Chen, Yu, Tang, Ji, Mu, Wu, et~al.]{chen2024matters}
Jiayu Chen, Chao Yu, Yuqing Xie, Feng Gao, Yinuo Chen, Shu'ang Yu, Wenhao Tang, Shilong Ji, Mo~Mu, Yi~Wu, et~al.
\newblock What matters in learning a zero-shot sim-to-real rl policy for quadrotor control? a comprehensive study.
\newblock \emph{arXiv preprint arXiv:2412.11764}, 2024.

\bibitem[Cheng et~al.(2024)Cheng, Kim, Song, Yang, Jin, Wang, and Hovakimyan]{cheng2024difftune}
Sheng Cheng, Minkyung Kim, Lin Song, Chengyu Yang, Yiquan Jin, Shenlong Wang, and Naira Hovakimyan.
\newblock Difftune: Auto-tuning through auto-differentiation.
\newblock \emph{IEEE Transactions on Robotics}, 2024.

\bibitem[Dionigi et~al.(2024)Dionigi, Costante, and Loianno]{dionigi2024power}
Alberto Dionigi, Gabriele Costante, and Giuseppe Loianno.
\newblock The power of input: Benchmarking zero-shot sim-to-real transfer of reinforcement learning control policies for quadrotor control.
\newblock \emph{arXiv preprint arXiv:2410.07686}, 2024.

\bibitem[Eschmann et~al.(2024)Eschmann, Albani, and Loianno]{eschmann2024learning}
Jonas Eschmann, Dario Albani, and Giuseppe Loianno.
\newblock Learning to fly in seconds.
\newblock \emph{IEEE Robotics and Automation Letters}, 2024.

\bibitem[Faessler et~al.(2018)Faessler, Franchi, and Scaramuzza]{faessler2018differential}
Matthias Faessler, Antonio Franchi, and Davide Scaramuzza.
\newblock Differential flatness of quadrotor dynamics subject to rotor drag for accurate tracking of high-speed trajectories.
\newblock \emph{IEEE Robotics and Automation Letters}, 3\penalty0 (2):\penalty0 620–626, April 2018.
\newblock ISSN 2377-3774.
\newblock \doi{10.1109/lra.2017.2776353}.
\newblock URL \url{http://dx.doi.org/10.1109/LRA.2017.2776353}.

\bibitem[F{\"o}rster(2015)]{forster2015system}
Julian F{\"o}rster.
\newblock System identification of the crazyflie 2.0 nano quadrocopter.
\newblock {B.S.} thesis, ETH Zurich, 2015.

\bibitem[Gupta et~al.(2023)Gupta, Chandak, Jordan, Thomas, and da~Silva]{gupta2023behavior}
Dhawal Gupta, Yash Chandak, Scott~M. Jordan, Philip~S. Thomas, and Bruno~Castro da~Silva.
\newblock Behavior alignment via reward function optimization, 2023.
\newblock URL \url{https://arxiv.org/abs/2310.19007}.

\bibitem[Huang et~al.(2023)Huang, Rana, Spitzer, Shi, and Boots]{huang2023datt}
Kevin Huang, Rwik Rana, Alexander Spitzer, Guanya Shi, and Byron Boots.
\newblock Datt: Deep adaptive trajectory tracking for quadrotor control.
\newblock In Jie Tan, Marc Toussaint, and Kourosh Darvish, editors, \emph{Proceedings of The 7th Conference on Robot Learning}, volume 229 of \emph{Proceedings of Machine Learning Research}, pages 326--340. PMLR, 06--09 Nov 2023.
\newblock URL \url{https://proceedings.mlr.press/v229/huang23a.html}.

\bibitem[Kaufmann et~al.(2022)Kaufmann, Bauersfeld, and Scaramuzza]{kaufmann2022benchmark}
Elia Kaufmann, Leonard Bauersfeld, and Davide Scaramuzza.
\newblock A benchmark comparison of learned control policies for agile quadrotor flight.
\newblock In \emph{2022 International Conference on Robotics and Automation (ICRA)}, pages 10504--10510. IEEE, 2022.

\bibitem[Kaufmann et~al.(2023)Kaufmann, Bauersfeld, Loquercio, M{\"u}ller, Koltun, and Scaramuzza]{kaufmann2023champion}
Elia Kaufmann, Leonard Bauersfeld, Antonio Loquercio, Matthias M{\"u}ller, Vladlen Koltun, and Davide Scaramuzza.
\newblock Champion-level drone racing using deep reinforcement learning.
\newblock \emph{Nature}, 620\penalty0 (7976):\penalty0 982--987, 2023.

\bibitem[Lee et~al.(2010)Lee, Leok, and McClamroch]{lee2010geometric}
Taeyoung Lee, Melvin Leok, and N~Harris McClamroch.
\newblock Geometric tracking control of a quadrotor uav on se (3).
\newblock In \emph{49th IEEE conference on decision and control (CDC)}, pages 5420--5425. IEEE, 2010.

\bibitem[Liang et~al.(2018)Liang, Makoviychuk, Handa, Chentanez, Macklin, and Fox]{liang2018gpu}
Jacky Liang, Viktor Makoviychuk, Ankur Handa, Nuttapong Chentanez, Miles Macklin, and Dieter Fox.
\newblock Gpu-accelerated robotic simulation for distributed reinforcement learning.
\newblock In \emph{Conference on Robot Learning}, pages 270--282. PMLR, 2018.

\bibitem[Loquercio et~al.(2022)Loquercio, Saviolo, and Scaramuzza]{loquercio2022autotune}
Antonio Loquercio, Alessandro Saviolo, and Davide Scaramuzza.
\newblock Autotune: Controller tuning for high-speed flight.
\newblock \emph{IEEE Robotics and Automation Letters}, 7\penalty0 (2):\penalty0 4432--4439, 2022.

\bibitem[Mellinger et~al.(2011)Mellinger, Lindsey, Shomin, and Kumar]{mellinger2011design}
Daniel Mellinger, Quentin Lindsey, Michael Shomin, and Vijay Kumar.
\newblock Design, modeling, estimation and control for aerial grasping and manipulation.
\newblock In \emph{2011 IEEE/RSJ International Conference on Intelligent Robots and Systems}, pages 2668--2673, 2011.
\newblock \doi{10.1109/IROS.2011.6094871}.

\bibitem[Mittal et~al.(2023)Mittal, Yu, Yu, Liu, Rudin, Hoeller, Yuan, Singh, Guo, Mazhar, Mandlekar, Babich, State, Hutter, and Garg]{mittal2023orbit}
Mayank Mittal, Calvin Yu, Qinxi Yu, Jingzhou Liu, Nikita Rudin, David Hoeller, Jia~Lin Yuan, Ritvik Singh, Yunrong Guo, Hammad Mazhar, Ajay Mandlekar, Buck Babich, Gavriel State, Marco Hutter, and Animesh Garg.
\newblock Orbit: A unified simulation framework for interactive robot learning environments.
\newblock \emph{IEEE Robotics and Automation Letters}, 8\penalty0 (6):\penalty0 3740--3747, 2023.
\newblock \doi{10.1109/LRA.2023.3270034}.

\bibitem[Molchanov et~al.(2019)Molchanov, Chen, H{\"o}nig, Preiss, Ayanian, and Sukhatme]{molchanov2019sim}
Artem Molchanov, Tao Chen, Wolfgang H{\"o}nig, James~A Preiss, Nora Ayanian, and Gaurav~S Sukhatme.
\newblock Sim-to-(multi)-real: Transfer of low-level robust control policies to multiple quadrotors.
\newblock In \emph{2019 IEEE/RSJ International Conference on Intelligent Robots and Systems (IROS)}, pages 59--66. IEEE, 2019.

\bibitem[Moreno-Valenzuela et~al.(2018)Moreno-Valenzuela, P{\'e}rez-Alcocer, Guerrero-Medina, and Dzul]{moreno2018nonlinear}
Javier Moreno-Valenzuela, Ricardo P{\'e}rez-Alcocer, Manuel Guerrero-Medina, and Alejandro Dzul.
\newblock Nonlinear pid-type controller for quadrotor trajectory tracking.
\newblock \emph{IEEE/ASME transactions on mechatronics}, 23\penalty0 (5):\penalty0 2436--2447, 2018.

\bibitem[Rudin et~al.(2022)Rudin, Hoeller, Reist, and Hutter]{rudin2022learning}
Nikita Rudin, David Hoeller, Philipp Reist, and Marco Hutter.
\newblock Learning to walk in minutes using massively parallel deep reinforcement learning.
\newblock In Aleksandra Faust, David Hsu, and Gerhard Neumann, editors, \emph{Proceedings of the 5th Conference on Robot Learning}, volume 164 of \emph{Proceedings of Machine Learning Research}, pages 91--100. PMLR, 08--11 Nov 2022.
\newblock URL \url{https://proceedings.mlr.press/v164/rudin22a.html}.

\bibitem[Schulman et~al.(2017)Schulman, Wolski, Dhariwal, Radford, and Klimov]{schulman2017proximal}
John Schulman, Filip Wolski, Prafulla Dhariwal, Alec Radford, and Oleg Klimov.
\newblock Proximal policy optimization algorithms.
\newblock \emph{arXiv preprint arXiv:1707.06347}, 2017.

\bibitem[Skalse et~al.(2022)Skalse, Howe, Krasheninnikov, and Krueger]{skalse2022defining}
Joar Skalse, Nikolaus H.~R. Howe, Dmitrii Krasheninnikov, and David Krueger.
\newblock Defining and characterizing reward hacking, 2022.
\newblock URL \url{https://arxiv.org/abs/2209.13085}.

\bibitem[Sun et~al.(2024)Sun, Romero, Foehn, Kaufmann, and Scaramuzza]{sun2024comparativestudynonlinearmpc}
Sihao Sun, Angel Romero, Philipp Foehn, Elia Kaufmann, and Davide Scaramuzza.
\newblock A comparative study of nonlinear mpc and differential-flatness-based control for quadrotor agile flight, 2024.
\newblock URL \url{https://arxiv.org/abs/2109.01365}.

\bibitem[Tal and Karaman(2021)]{tal2021accurate}
Ezra Tal and Sertac Karaman.
\newblock Accurate tracking of aggressive quadrotor trajectories using incremental nonlinear dynamic inversion and differential flatness.
\newblock \emph{IEEE Transactions on Control Systems Technology}, 29\penalty0 (3):\penalty0 1203--1218, 2021.
\newblock \doi{10.1109/TCST.2020.3001117}.

\bibitem[Wang et~al.(2016)Wang, Yuan, and Zhu]{wang2016automatic}
Wufan Wang, Xiaming Yuan, and Jihong Zhu.
\newblock Automatic pid tuning via differential evolution for quadrotor uavs trajectory tracking.
\newblock In \emph{2016 IEEE Symposium Series on Computational Intelligence (SSCI)}, pages 1--8. IEEE, 2016.

\bibitem[Welde et~al.(2021)Welde, Paulos, and Kumar]{welde2021dynamically}
Jake Welde, James Paulos, and Vijay Kumar.
\newblock Dynamically feasible task space planning for underactuated aerial manipulators.
\newblock \emph{IEEE Robotics and Automation Letters}, 6\penalty0 (2):\penalty0 3232--3239, 2021.
\newblock \doi{10.1109/LRA.2021.3051572}.

\bibitem[Welde et~al.(2024)Welde, Rao, Kunapuli, Jayaraman, and Kumar]{welde2024leveraging}
Jake Welde, Nishanth Rao, Pratik Kunapuli, Dinesh Jayaraman, and Vijay Kumar.
\newblock Leveraging symmetry to accelerate learning of trajectory tracking controllers for free-flying robotic systems.
\newblock \emph{arXiv preprint arXiv:2409.11238}, 2024.

\bibitem[Zhang et~al.(2023)Zhang, Loquercio, Wu, Kumar, Malik, and Mueller]{zhang2023learning}
Dingqi Zhang, Antonio Loquercio, Xiangyu Wu, Ashish Kumar, Jitendra Malik, and Mark~W Mueller.
\newblock Learning a single near-hover position controller for vastly different quadcopters.
\newblock In \emph{2023 IEEE International Conference on Robotics and Automation (ICRA)}, pages 1263--1269. IEEE, 2023.

\bibitem[Zhang et~al.(2025)Zhang, Rivera, Tyni, and Nguyen]{zhang2025airpilot}
Junyang Zhang, Cristian Emanuel~Ocampo Rivera, Kyle Tyni, and Steven Nguyen.
\newblock Airpilot: Interpretable ppo-based drl auto-tuned nonlinear pid drone controller for robust autonomous flights, 2025.
\newblock URL \url{https://arxiv.org/abs/2404.00204}.

\bibitem[Zhu et~al.(2015)Zhu, Liu, Guo, and Xu]{zhu2015gradient}
Jiangcheng Zhu, Endong Liu, Shan Guo, and Chao Xu.
\newblock A gradient optimization based pid tuning approach on quadrotor.
\newblock In \emph{The 27th Chinese Control and Decision Conference (2015 CCDC)}, pages 1588--1593. IEEE, 2015.

\end{thebibliography}

\clearpage
\nobalance
\begin{appendices}

\section{Task Specifications}
\label{sec:appendix_task}
We enumerate the specific ranges of values used in the various tasks in terms of randomized parameters for the trajectory and initialization in Table \ref{tab:appendix_task_specification}. 

\begin{table}[h]
    \centering
    \ra{1.2}
    \begin{tabular}{@{}l l l l l l @{}}
        \toprule
        & & \multicolumn{4}{c}{Lissajous Parameters} \\
        \cmidrule(lr){3-6} 
        Task  & State & $A$ & $\omega$ & $\phi$ & $\delta$  \\
        \midrule
        \multirow{4}{*}{Hover} & $x$ & [0.0, 0.0] & [-2.0, 2.0] & [-$\pi$, $\pi$] & [-2.0, 2.0]   \\ 
         & $y$ & [0.0, 0.0] & [-2.0, 2.0] & [-$\pi$, $\pi$] & [-2.0, 2.0] \\
         & $z$ & [0.0, 0.0] & [-2.0, 2.0] & [-$\pi$, $\pi$] & [-2.0, 2.0] \\
         & $\psi$ & [0.0, 0.0] & [-2.0, 2.0] & [-$\pi$, $\pi$] & [-$\pi$, $\pi$] \\
         \midrule
        \multirow{4}{*}{Lissajous Tracking} & $x$ & [-2.0, 2.0] & [-3.0, 3.0] & [-$\pi$, $\pi$] & [-2.0, 2.0]  \\ 
         & $y$ & [-2.0, 2.0] & [-3.0, 3.0] & [-$\pi$, $\pi$] & [-2.0, 2.0] \\
         & $z$ & [-2.0, 2.0] & [-3.0, 3.0] & [-$\pi$, $\pi$] & [-2.0, 2.0] \\
         & $\psi$ & [-2.0, 2.0] & [-2.0, 2.0] & [-$\pi$, $\pi$] & [-$\pi$, $\pi$]\\
        \bottomrule        
    \end{tabular}
    \caption{\textbf{Randomization Ranges for Tasks.} Ranges used for both optimization and evaluation in Hover and Lissajous Tracking tasks, listed in terms of randomized Lissajous parameters.}
    \label{tab:appendix_task_specification}
\end{table}

\section{GC Implementation Details}
\label{sec:appendix_gc_implementation}
The geometric controller (GC) observes the state of the system, commonly modeled as the center-of-mass (COM) of the quadrotor in the world frame. This information is the position $\boldsymbol{p}$, orientation $\boldsymbol{R}$, linear velocity $\boldsymbol{v}$, and angular velocity $\boldsymbol{\omega}$. Conditioned on some goal, specified by the desired position $\boldsymbol{p}_{d}$ and desired yaw $\psi_d$ as well as $4^{\textrm{th}}$ order derivatives of the desired position and $2^{\textrm{nd}}$ order derivatives of the desired yaw, the GC computes a desired control by using a cascaded PD control strategy for the position and attitude separately. 
The GC first computes a desired acceleration from the position and derivatives of the reference trajectory using a PD-control structure:
\begin{equation}
    \ddot{\boldsymbol{p}}_{des} = -K_p(\boldsymbol{p} - \boldsymbol{p}_d) -K_v(\boldsymbol{v} - \boldsymbol{v}_d) - mg\boldsymbol{z}_{\mathcal{W}} + \ddot{\boldsymbol{p}}_{d},
\end{equation}
Then, from the desired acceleration and desired yaw, the desired orientation can be computed according to \citep[eq. ~(13-14)]{welde2021dynamically}:
\begin{equation}
    \label{eq:desired_attitude}
    \boldsymbol{R}_{des} = H_1(\psi_d)H_2(\frac{\ddot{\boldsymbol{p}}_{des}}{\norm{\ddot{\boldsymbol{p}}_{des}}}),
\end{equation}
and the attitude control loop uses a second PD-control structure with feed-forward. We obtain $\boldsymbol{\omega}_d$ by differentiating \eqref{eq:desired_attitude}:
\begin{equation}
    \label{eq:gc_att_loop}
    \boldsymbol{e_R} = \frac{1}{2}(\boldsymbol{R}^{T}_{des} \boldsymbol{R} - \boldsymbol{R}^T \boldsymbol{R}_{des})^{\vee},
\end{equation}
\begin{equation}
\begin{multlined}
    \boldsymbol{\dot\omega}_{des} = -K_R(\boldsymbol{e_R}) -K_{\omega}(\boldsymbol{\omega} - \boldsymbol{\omega}_d) - \\ (\hat{\boldsymbol{\omega}}\boldsymbol{R}^T\boldsymbol{R}_{des}\boldsymbol{\omega}_d - \boldsymbol{R}^T\boldsymbol{R}_{des}\boldsymbol{\dot\omega}_d).
\end{multlined}
\end{equation}
Finally, the control input can be computed according to \cite{lee2010geometric}:
\begin{equation}
    f_T = m\ddot{\boldsymbol{p}}_{des} \cdot \boldsymbol{R}\boldsymbol{z}_{\mathcal{V}},  \label{eq:gc_f}
\end{equation}
\begin{equation}
    \boldsymbol{M} = \boldsymbol{J}(\boldsymbol{\dot\omega}_{des}) + \boldsymbol{\omega} \times \boldsymbol{J} \boldsymbol{\omega}. \label{eq:gc_M}
\end{equation}

\section{RL Implementation Details}
\label{sec:appendix_rl_implementation}
The RL policy is parameterized by a neural network, representing a wide range of hypothesis classes for the structure of the policy. For both the actor and critic, we use a network with 3 layers, each with 256 nodes, activated by ELU, totaling 275,717 learnable parameters. To optimize this network, we use PPO \cite{schulman2017proximal} from the RSL-RL library \cite{rudin2022learning}, and perform 750 total updates on rollouts from 4096 simultaneous agents of 64 timesteps each. We found annealing the position tolerance $\delta_p$ key to enabling agile behavior while still converging to the goal without large steady-state error. We reduced $\delta_{p}$ by half every 50M timesteps, going from 0.8 at the beginning of training and ending at 0.1 as seen in Table \ref{tab:hyperparameters}.  We define our observation as follows:
\begin{equation}
     o_t = \begin{bmatrix}
        \customVector{e}{B}{}{p} \\
        \customVector{e}{B}{}{R} \\
        \customVector{g}{B}{}{} \\
        \customVector{e}{B}{}{v} \\
        \customVector{e}{B}{}{\omega}                
    \end{bmatrix} = \begin{bmatrix}
        \customVector{R}{B}{W}{}(\customVector{p}{W}{B}{} - \customVector{p}{W}{B}{d}) \\
        (\customVector{R}{W}{B}{})^T \customVector{R}{W}{B}{d} \\
        \customVector{R}{B}{W}{}(g \boldsymbol{z}_{\mathcal{W}}) \\
        \customVector{R}{B}{W}{}(\customVector{v}{W}{B}{} - \customVector{v}{W}{B}{d}) \\
        \customVector{R}{B}{W}{}(\customVector{\omega}{W}{B}{} - \customVector{\omega}{W}{B}{d}),    
    \end{bmatrix}
\end{equation}
where $\customVector{R}{B}{W}{}$ is the rotation from the world frame $\mathcal{W}$ to the body $\mathcal{B}$, the position of the body in the world frame is $\customVector{p}{W}{B}{}$, the linear velocity of the body in the world frame is $\customVector{v}{W}{B}{}$, and the angular velocity in the world frame is $\customVector{\omega}{W}{B}{}$. The observation errors are computed relative to the desired position of the body in the world frame $\customVector{p}{W}{B}{d}$, desired rotation from the body frame to the world frame $\customVector{R}{W}{B}{d}$, desired linear velocity of the body in the world frame $\customVector{v}{W}{B}{d}$, and desired angular velocity of the body in world frame $\customVector{\omega}{W}{B}{d}$. 

\begin{table*}[t]
    \centering
    \begin{tabular}{@{}llllllll@{}}
         \toprule
         & \multicolumn{3}{c}{Lissajous Tracking} & \multicolumn{3}{c}{Hover} \\
         \cmidrule(lr){2-4} \cmidrule(lr){5-7}
         Model & Reward & Position RMSE & Yaw RMSE &  Reward & Position RMSE & Yaw RMSE \\
        \midrule
        RL-Opt-Liss.-FF  &  13.621  $\pm$  0.816  &  0.117  $\pm$  0.051  &  0.487  $\pm$  0.265  &  12.409  $\pm$  1.214  &  0.423  $\pm$  0.178  &  0.606  $\pm$  0.255 \\
        GC-Man-Lissajous-FF  &  13.494  $\pm$  1.477  &  0.167  $\pm$  0.144  &  0.44  $\pm$  0.283 \\
        GC-Opt-Lissajous-FF  &  13.792  $\pm$  1.278  &  0.136  $\pm$  0.099  &  0.404  $\pm$  0.294 &  13.474  $\pm$  0.692  &  0.372  $\pm$  0.121  &  0.451  $\pm$  0.21 \\
        RL-Opt-Hover-FF  &  13.494  $\pm$  0.863  &  0.172  $\pm$  0.079  &  0.659  $\pm$  0.353 &  13.918  $\pm$  0.335  &  0.327  $\pm$  0.104  &  0.509  $\pm$  0.285 \\
        GC-Opt-Hover-FF  &  13.695  $\pm$  1.261  &  0.153  $\pm$  0.118  &  0.416  $\pm$  0.292  &  13.72  $\pm$  0.477  &  0.383  $\pm$  0.117  &  0.333  $\pm$  0.172\\
        RL-Opt-Lissajous-None  &  11.559  $\pm$  2.172  &  0.406  $\pm$  0.191  &  0.354  $\pm$  0.142 \\
        GC-Opt-Lissajous-None  &  11.937  $\pm$  2.19  &  0.318  $\pm$  0.29  &  0.629  $\pm$  0.343 \\
        GC-Opt-Lissajous-PID  &  11.524  $\pm$  2.458  &  0.35  $\pm$  0.235  &  0.627  $\pm$  0.385 \\
        GC-Man-Hover-FF  &&&&  12.953  $\pm$  1.022  &  0.41  $\pm$  0.153  &  0.501  $\pm$  0.193 \\
        RL-Opt-Hover-None  &&&&  13.918  $\pm$  0.335  &  0.327  $\pm$  0.104  &  0.509  $\pm$  0.285 \\
        GC-Opt-Hover-None  &&&&  13.756  $\pm$  0.449  &  0.355  $\pm$  0.116  &  0.421  $\pm$  0.139 \\
        GC-Opt-Hover-PID  &&&&  9.955  $\pm$  3.009  &  0.631  $\pm$  0.351  &  0.854  $\pm$  0.265 \\
        \bottomrule
    \end{tabular}
    \caption{Achieved rewards, position RMSE, and yaw RMSE for controller configurations tested in Lissajous Tracking and Hover tasks. Results are shown as mean with standard deviation over 1000 trials per task.}
    \label{tab:model_ablation_full}
\end{table*}

\section{Motor Dynamics}
\label{sec:appendix_motor_dynamics}
Motor dynamics are implemented by using a first-order delay, as is common for modeling quadrotor motor dynamics \cite{forster2015system}. 
From a desired control input $f_T$, and $\boldsymbol{M}$, we can compute the desired motor speeds $\boldsymbol{\Omega}_d$ by solving the following equations:
\begin{equation}
    \boldsymbol{\Omega}_d = \begin{bmatrix}
        \Omega_{1d} \enspace \Omega_{2d} \enspace \Omega_{3d} \enspace \Omega_{4d}
    \end{bmatrix}^T
\end{equation}
\begin{equation}
    f_t = k_t * [1, 1, 1, 1] * \boldsymbol{\Omega}_d^2
\end{equation}
\begin{equation}
    \boldsymbol{M} = \begin{bmatrix}
        0, -l\cdot k_t, 0, l\cdot k_t\\
        l\cdot k_t, 0, -l\cdot k_t, 0\\
        -k_m, k_m, -k_m, k_m
    \end{bmatrix}  *\boldsymbol{\Omega}_d^2,
\end{equation}
where $k_t$ is the motor-thrust coefficient, $k_m$ is the motor-torque coefficient, and $l$ is the arm-length from the center-of-mass to the motor. 
Finally, we apply the motor delay with $\tau_m$ representing the motor delay:
\begin{equation}
    \boldsymbol{\dot\Omega} = \frac{1}{\tau_m}(\boldsymbol{\Omega}_d - \boldsymbol{\Omega}).
\end{equation}

\section{Supplement: Trajectory Tracking Results}
A full presentation of trajectory tracking results performed accompanying section V-C is presented here. Each model is evaluated on 1000 rollouts, and the resulting achieved reward, position RMSE, and yaw RMSE is shown for both the Lissajous Tracking and Hover tasks in Table \ref{tab:model_ablation_full}. 

\end{appendices}
\end{document}